\documentclass{article}
\pdfoutput=1
\usepackage[utf8]{inputenc}
\usepackage[T1]{fontenc}
\usepackage{microtype}
\usepackage{graphicx}
\usepackage{subcaption}
\usepackage{booktabs}
\PassOptionsToPackage{hyphens}{url}
\usepackage[hyperfootnotes=false]{hyperref}

\usepackage[preprint]{icml2026}
\hypersetup{pdftitle={Measuring and Teaching Language Model Agents to Act on Delayed Consequences},pdfsubject={Preprint}}

\makeatletter
\def\toptitlebar{\nointerlineskip\vbox to 0pt{\vss\hbox{\includegraphics[height=10pt]{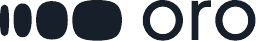}}\kern5pt}\hrule height1pt \vskip .25in}
\makeatother

\usepackage{amsmath}
\usepackage{amssymb}
\usepackage{mathtools}
\usepackage{amsthm}

\newcommand{\negusd}[1]{\mbox{\ensuremath{-}\$#1}}
\newcommand{\negnum}[1]{\mbox{\ensuremath{-}#1}}

\DeclareUnicodeCharacter{2248}{\ensuremath{\approx}}
\DeclareUnicodeCharacter{2192}{\ensuremath{\rightarrow}}
\DeclareUnicodeCharacter{2212}{\ensuremath{-}}

\newenvironment{tracequote}{%
  \par\vspace{2.5pt plus 2.5pt minus 1pt}\noindent\hspace*{0.7em}%
  \begin{minipage}{0.94\linewidth}%
  \footnotesize\ttfamily\raggedright
}{%
  \end{minipage}\par\vspace{2.5pt plus 2.5pt minus 1pt}
}

\icmltitlerunning{Measuring and Teaching Language Model Agents to Act on Delayed Consequences}

\begin{document}

\twocolumn[
  \icmltitle{Measuring and Teaching Language Model Agents\texorpdfstring{\\}{ }to Act on Delayed Consequences}

  \begin{icmlauthorlist}
    \icmlauthor{Shardul Bansal}{oro}
    \icmlauthor{Seth Schilbe}{oro}
    \icmlauthor{Jarrod Barnes}{oro}
  \end{icmlauthorlist}

  \icmlaffiliation{oro}{Oro AI, San Francisco, CA, USA}

  \icmlcorrespondingauthor{Shardul Bansal}{Shardul@oroagents.com}

  \icmlkeywords{Language Model Agents, Reinforcement Learning, Delayed Consequences, Benchmarks, Operations Research}

  \vskip 0.20in
]

\printAffiliationsAndNotice{\raggedright Benchmark report at \href{https://oroagents.com/benchmarks/smb-bench}{\texttt{https://\allowbreak oroagents.\allowbreak com/\allowbreak benchmarks/\allowbreak smb-\allowbreak bench}} and public dataset at \href{https://huggingface.co/datasets/oro-agents/smb-bench}{\texttt{https://\allowbreak huggingface.\allowbreak co/\allowbreak datasets/\allowbreak oro-\allowbreak agents/\allowbreak smb-\allowbreak bench}}.}

\begin{abstract}
Language model agents are increasingly asked to run ongoing multi-day operations rather than finish isolated single-turn tasks. In daily operations, today's choices reshape later observations and cash constraints because empty shelves hide lost walk-in demand, short-shipped customers quietly shrink future orders weeks later, and emergency buying drains working capital. We introduce SMB Bench, a calibrated 90-day wholesale simulator with 18 causally paired tasks and an 859-run frontier trace dataset, to ask what experience and learning signal make an agent better at noticing, attributing, and acting on the delayed consequences of its own decisions in unseen businesses. Across 16 frontier models, only 3 outperform a fixed buying rule; models often record a relevant diagnosis in their notes without changing later purchases in time, showing that diagnosing a problem, changing behavior, and improving a business are separate achievements. To study how experience improves those failures in an open model (Qwen3.8-27B), we used training-business records to identify operational rules and archived frontier quarters as starting states to continue from rather than transcripts to copy. On an unseen test business under identical evaluation instructions, the trained policy followed customer-protection commitments more consistently and reduced severe-loss quarters, while the additional profit benefit remained uncertain. Carrying those commitments through the quarter exposed the cost of fulfilling them, as early-quarter stocking habits left the buyer relying on emergency supply and a deteriorating supplier. The results separate recognizing a delayed effect, acting on it consistently, and coordinating surrounding decisions so that the response improves profit. SMB Bench connects these observations to a learning method by turning archived decisions and their consequences into reusable training experience.
\end{abstract}

\section{Introduction}
\label{sec-intro}

\begin{figure*}[t]
  \centering
  \includegraphics[width=\textwidth]{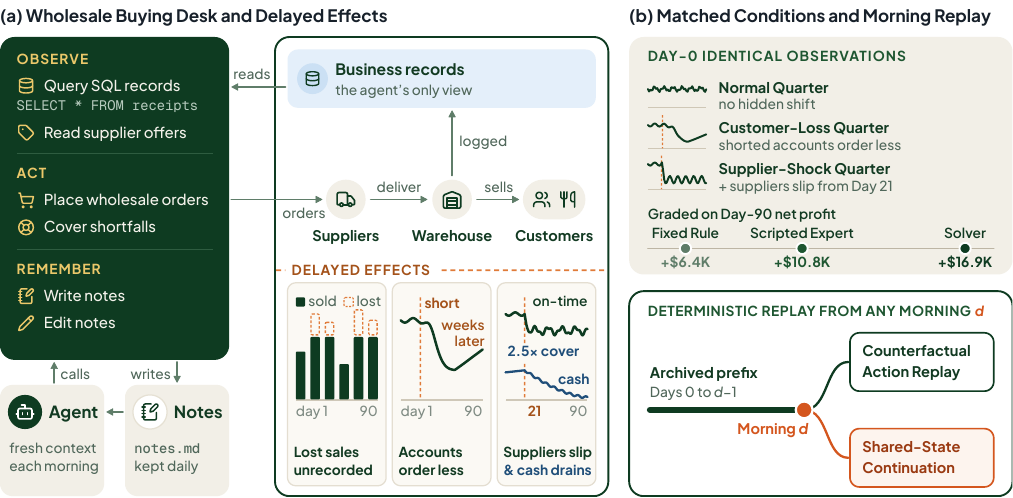}
  \caption{The SMB Bench environment, matched conditions, and deterministic state replay. (a) Across 90 mornings, the buyer queries SQL records and daily supplier quotes, places wholesale case orders, sets customer shortfall protection, and updates persistent notes. Decisions reshape later evidence and cash through hidden walk-in demand during stockouts, multi-week customer order reductions after shortfalls, and 2.5$\times$ emergency-cover cash drain. (b) Every business runs under three conditions with identical buyer-visible Day-0 observations (normal, customer-loss, and Day-21 supplier-shock quarters) and supports deterministic replay from any committed morning to compare alternative actions or continue archived runs.}
  \label{fig-overview}
\end{figure*}

In ongoing commercial operations, daily decisions reshape both the evidence an agent later observes and the working capital it retains. When a buyer lets shelves run empty, unserved walk-in demand goes unrecorded in sales logs \cite{huh2009censored,besbes2013censoring}; when an account delivery falls short, the customer reduces subsequent orders weeks after the original shortfall; and when emergency local purchases cover missing inventory, immediate service is preserved at the expense of cash needed for future stock. Assigning credit across weeks of delayed feedback has challenged AI since \citet{minsky1961steps}, because the agent must interpret observations partly generated by its own prior choices and adjust behavior before the full cost materializes \cite{polanyi1966tacit}.

Recent work has begun to measure sustained operation, adaptation, and learning from interaction \cite{silver2025experience}. Commercial and multi-day benchmarks expose planning, diagnostics, and coordinated business effects \cite{backlund2025vending,chen2026ceobench,raghuvanshi2026storebench}, behavioral audits separate collecting evidence or stating a diagnosis from acting on it \cite{zhang2026retailbench,deb2026stocktake,li2026genebench}, and stateful-versus-stateless evaluations separate gains from accumulated experience from starting competence \cite{asawa2026clbench}. Training methods likewise use interaction-generated future states and continuations from restored intermediate states to improve multi-turn agents \cite{ross2014aggrevate,zhang2025earlyexperience,zhang2026crps,li2026crr,liu2026sprout}.

These developments motivate a common question, what experience and learning signal make an agent better at noticing, attributing, and acting on the delayed consequences of its own decisions in unseen businesses? Our working hypothesis is that useful experience preserves the link between a decision, the state in which it was made, and its later consequences; making relationships in those records explicit can guide better choices, while practicing from recorded states can make those choices more consistent. Testing this hypothesis requires measuring what an agent records, what it does next, and whether the response improves business profit, three achievements that prove separable throughout our study.

To investigate all three in one system, we build SMB Bench \cite{oro2026smbbench}, a calibrated 90-day perishable wholesale environment \cite{wang2025freshretailnet} (Figure~\ref{fig-overview}; \href{https://oroagents.com/benchmarks/smb-bench}{benchmark report} and \href{https://huggingface.co/datasets/oro-agents/smb-bench}{public trace dataset}). Across six businesses and 18 tasks, each store runs under normal, customer-loss, and supplier-shock conditions with identical buyer-visible Day-0 observations alongside non-LLM reference policies and deterministic morning replay (Section~\ref{sec-environment}), letting us test whether behavior changes specifically when decisions carry delayed penalties and whether missed actions would have helped (Section~\ref{sec-measures}). Across 859 scored runs of 16 frontier models, only 3 outperform a Fixed Buying Rule (+\$6,412 versus +\$10,795 Scripted Expert). In the nine-model behavioral cohort, agents stocked below 60\% of their own written forecasts in 78.7\% (107 of 136) of audited cases; across quarters with reactive accounts, 21.4\% contained a written link between falling orders and prior shortfalls, and in 3.7\% of reactive-account quarters the agent both recorded that link and extended protection to all accounts within three mornings of the note (Section~\ref{sec-frontier}).

Using SMB Bench's replay on Qwen3.8-27B, we extract operational rules from training-business records, refine prompt and optimizer choices on a validation business, and train on continuations from archived frontier mornings rather than copied transcripts (Section~\ref{sec-learning}). On a sealed test business, the final system finishes ahead of 8 of 16 leaderboard models (\negusd{4,495} versus +\$1,969 for GPT-6 Astra); holding the unseen evaluation instructions fixed, the trained policy followed customer-protection commitments more consistently, cutting protection reversals from 5.8 to 1.0 per quarter and quarters losing over \$10,000 from 25\% to 8\% while measured recognition of stockout censoring changed little (84.7\% to 85.6\%) and the additional profit difference remained uncertain (+\$1,489 $\pm$ \$1,133 standard error). Consistently protecting customers in turn exposed an upstream coordination problem across inventory buffers, supplier lead times, and cash (Section~\ref{sec-coordination}), clarifying what autonomous experience-learning agents must coordinate next \cite{lin2026orbit,shalev2026reflector}.
\begin{itemize}\setlength{\topsep}{4pt plus 4pt minus 1pt}\setlength{\itemsep}{2pt plus 2pt}\setlength{\parskip}{0pt}
  \item We introduce \emph{SMB Bench}, a calibrated 90-day wholesale environment pairing matched Day-0 normal and consequence conditions with non-LLM reference policies and deterministic replay from any morning.
  \item We release an \emph{859-run frontier evaluation and trace dataset} across 16 models, connecting controlled cover trajectories, written-forecast audits, and restock replays to separate recorded diagnosis from consistent operational action.
  \item We turn archived runs into \emph{reusable training experience} for Qwen3.8-27B, testing what changes on a sealed business under fixed evaluation instructions and diagnosing the supply and cash limits that consistent customer protection exposes.
\end{itemize}

\section{SMB Bench}
\label{sec-environment}

To study delayed consequences cleanly, an environment must couple daily choices across weeks while keeping every intermediate state reproducible. SMB Bench places one model on the produce buying desk of a wholesaler over a 90-day quarter ($d = 0, \dots, 89$). Each store sells 18 to 20 perishable items to walk-in shoppers at shelf or markdown prices and to six restaurant and catering accounts at locked order-day prices. Every morning opens at 5 a.m.\ as a fresh conversation showing a status dashboard, the buyer prompt, and the model's persistent notebook (\texttt{notes.md}, up to 16 files and 32,000 bytes). With up to 60 tool calls per morning, the buyer queries DuckDB SQL tables (\texttt{query\_\allowbreak records}, \texttt{list\_\allowbreak tables}, \texttt{describe\_\allowbreak table}), inspects daily supplier quotes (\texttt{list\_\allowbreak offers}), places case orders (\texttt{place\_\allowbreak purchase\_\allowbreak order}), sets per-item or wildcard (\texttt{*}) shortfall protection (\texttt{set\_\allowbreak shortfall\_\allowbreak cover}), and updates notes before calling \texttt{advance\_\allowbreak day} alone. Fixed simulator rules run the rest of the day, receiving trucks before store open, filling account lines and automatic 6 a.m.\ local cover at $\sim$2.54$\times$ wholesale quotes, serving 16 hours of walk-in sales, stepping down walk-in prices past markdown age, discarding expired lots, and settling bills.

Episodes are graded once after Day 89 on the net change in business value. Purchases remain inventory assets until sold or written off, so the grader's accounting bridge decomposes that change as
\begin{align}
\mathrm{Profit} &= V_{\mathrm{close}} - V_{\mathrm{open}} = R - C_{\mathrm{cogs}} - C_{\mathrm{waste}} \nonumber \\
&\quad - C_{\mathrm{short}} - C_{\mathrm{ops}} - C_{\mathrm{lost}}, \label{eq-profit}
\end{align}
where $R$ is walk-in and account revenue net of customer return refunds, $C_{\mathrm{cogs}}$ is the wholesale and 6 a.m.\ local purchase cost of sold units, $C_{\mathrm{waste}}$ writes off expired, culled, and inspection-rejected lots plus closing stock that projected demand cannot sell before expiry, $C_{\mathrm{short}}$ is the 5\% compensation on first due-day shorted account value, and $C_{\mathrm{ops}}$ covers dispatch, delivery, receiving, refusal, disposal, storage, interest, and overhead fees ($0.1647\times$ baseline revenue in overhead, putting the reference buyer near a 3\% net margin) and pending refund and claim reserves, net of supplier claim recoveries. $C_{\mathrm{lost}}$ deducts the 28-day closing customer-demand reserve at 30\% of price when prior shorts have shrunk account orders. Each unit bought therefore enters profit exactly once, through $C_{\mathrm{cogs}}$ when sold or $C_{\mathrm{waste}}$ when lost, and the closing write-down and $C_{\mathrm{lost}}$ block quarter-end gaming.

Realism comes from public datasets checked against wholesaler workflows with industry partner GetASAP, using zero private company records. Hourly walk-in demand and stockout censoring patterns are fit from \texttt{Fresh\-Retail\-Net-\allowbreak 50K} \cite{wang2025freshretailnet}, price and freight spreads draw on \texttt{USDA/\allowbreak ERS} reports, and cooler groups (\texttt{cold\_\allowbreak dry}, \texttt{cold\_\allowbreak humid}, \texttt{cool}, \texttt{warm}) and physical lifetimes (7 to 240 days) follow \texttt{UC Davis} postharvest guides. Wholesaler consultations grounded two cross-product couplings, supplier minimum-weight thresholds for dispatch fees and shared working capital (spendable cash plus unused credit minus open payables and open purchase orders, tracked as \texttt{available\_\allowbreak to\_\allowbreak commit}), funding both wholesale orders and 6 a.m.\ local cover.

Before freezing the benchmark, we screened 42 candidate tasks across 168 worlds so scores isolate delayed adaptation rather than Day-0 clues (Table~\ref{tab-admission} in Appendix~\ref{app-benchmark}). Every world matched its twin control byte-for-byte on all buyer-visible Day-0 surfaces and passed an independent accounting audit. Admission required $\ge 0.25\%$ baseline-revenue damage to an ignoring buyer in at least 4 of 6 stores, a feasible remedy, and $0/0/0$ false positives on blind-habit, generous-default, and twin-chance checks. Four mechanisms fell below the 4-store materiality bar, leaving 18 tasks across six businesses, 6 normal quarters, 6 customer-loss quarters with reactive accounts (shortfalls shrink future account orders, costing two-thirds of volume after three weeks of 10\% shorts), and 6 supplier-shock quarters (reactive accounts plus the cheapest supplier's late and cancel rates jump on orders placed from Day 21 onward).

Every task is anchored by three non-LLM controllers. Each morning, the \emph{Fixed Buying Rule} (+\$6,412 mean across 18 tasks, +\$10,051 normal, +\$4,592 changed) reads the past 14 days of censoring-adjusted walk-in and account sales, booked account lines, on-hand and in-transit stock, and current supplier quotes; it orders enough from the cheapest quote to cover due account orders plus delivery lead time and a 1.5-day buffer, covers shortfalls locally only within 5\% of contract price, keeps no notes, and never alters its rule when conditions shift. The \emph{Scripted Expert Buyer} (+\$10,795) encodes wholesaler practices on buyer-visible data, consolidating supplier orders to save dispatch fees, covering all accounts from Day 0, and holding lead-time buffers. The \emph{Perfect-Foresight Solver} (+\$16,923) replans daily with oracle future knowledge. Paired with deterministic replay from any committed morning, these matched conditions and reference policies provide states and outcomes that can be reused for counterfactual testing and training.

\section{Evaluation and Behavioral Measures}
\label{sec-measures}

To distinguish starting policies, generic mid-quarter drift, and responses to delayed consequences, our benchmark analysis combines full-leaderboard scoring with pre-specified trace audits and replay tests (Table~\ref{tab-cohorts}). All 16 models ran the unedited benchmark prompt across all 18 tasks (6 businesses $\times$ 3 conditions, 3 repeats per cell), yielding 859 valid scored quarters of 864 attempted (five runs ended on provider execution errors and remain unscored, while 20 infrastructure-interrupted runs resumed deterministically from their last committed morning). Because store sizes span \$91K to \$340K in 90-day baseline sales potential, task comparisons report both dollar profit and baseline-revenue-normalized margin points over the Fixed Buying Rule in the identical world (Table~\ref{tab-full-leaderboard}).

\begin{table*}[t]
  \caption{Populations and selection rules for each benchmark evaluation and behavioral analysis. Quarters with reactive accounts include both customer-loss and supplier-shock tasks (9 models $\times$ 12 tasks), whereas the cover trajectory uses customer-loss quarters alone (6 tasks, 18 runs per model).}
  \label{tab-cohorts}
  \vskip 0.04in
  \centering
  \footnotesize\setlength{\tabcolsep}{7pt}
  \begin{tabular}{lll}
    \toprule
    Measure & Population & Qualification and Scoring Rule \\
    \midrule
    Benchmark standings & 16 models, 859 runs & 18 tasks $\times$ 3 repeats; net value vs.\ Fixed Rule \\
    Volume vs.\ profit & 9 models, 485 runs & Behavioral cohort; kg bought vs.\ Fixed Rule \\
    Written forecast audit & 136 product cases & Numeric forecast in notes vs.\ stocked units \\
    Post-stockout replay & 50 restock-note cases & Astra (26), Sol (24) after $\ge 25\%$ OOS weeks \\
    Diagnosis funnel & 323 quarters with reactive accounts & Loss and shock tasks; written link, 3-day cover \\
    Course-change rate & 323 vs.\ 162 normal & Broad cover shift after Day 1 \\
    Cover trajectory & 9 models, 18 vs.\ 18 & Customer-loss quarters only; rise from Day 14 to 88 \\
    \bottomrule
  \end{tabular}
\end{table*}

Detailed behavioral analysis covers the nine models in the original benchmark release (485 scored runs across GPT-6 Astra, Claude Fable 5.1, GPT-6.1 Sol, Gemini 4 Argon, Claude Opus 5.5, DeepSeek V4.1 Flash, Grok 4.7, GLM 5.3, and Muse Wonder 1.0), while leaderboard comparisons include all 16 evaluated models. Within that cohort, forecast execution audits every product-morning where a model recorded an explicit numeric demand estimate in its notes (136 cases), comparing stocked quantity against its stated forecast. Counterfactual restock replays examine weeks with at least 25\% out-of-stock hours where GPT-6 Astra (26 cases) or GPT-6.1 Sol (24 cases) wrote a note calling for higher stock, replaying the episode with those noted orders placed.

For delayed customer loss, we audit all 323 scored quarters with reactive accounts, which include both the customer-loss and supplier-shock tasks (9 models $\times$ 12 tasks), alongside 162 normal-quarter controls of the same businesses. A run records \emph{written attribution} when notes or reasoning explicitly link falling account orders to the buyer's prior delivery shortfalls, and \emph{prompt action} when standing shortfall cover is extended to all six accounts within three mornings of that note. Cover rules already enabled during Day~0--1 initial setup are excluded from mid-quarter course changes. Finally, to test controlled adaptation over time, we track each model's daily share of account-product lines protected by shortfall cover across Days 0 to 88 (stopping before Day 89, when newly set cover can no longer protect an order) and compare the Day-14-to-Day-88 rise in customer-loss quarters alone (6 tasks, 18 runs per model, excluding supplier-shock quarters) against the same model's normal quarters via 10,000-resample task-and-run bootstrap intervals.

\section{Frontier Results}
\label{sec-frontier}

\begin{figure*}[t]
  \centering
  \includegraphics[width=\textwidth]{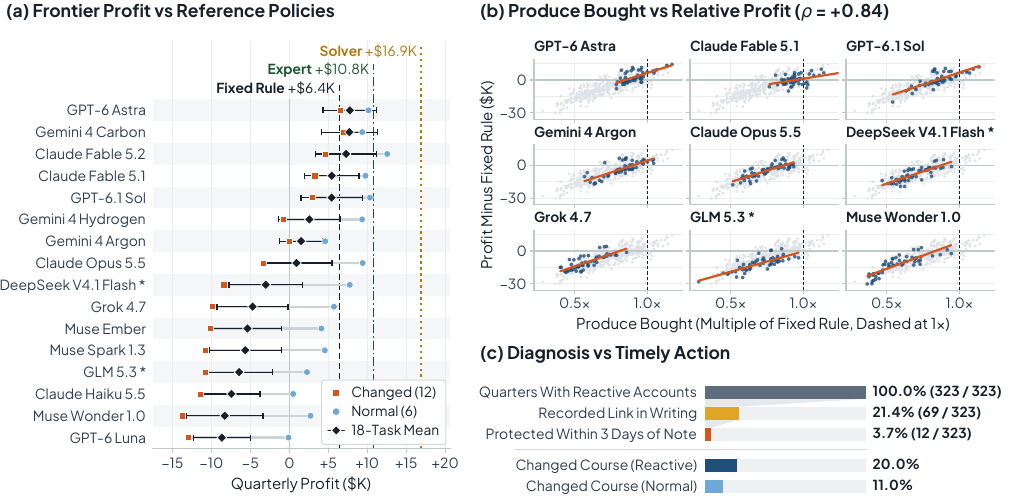}
  \caption{Frontier benchmark standings and behavioral diagnostics. (a) Mean profit across 16 models and 859 scored runs on all 18 tasks (diamonds, 95\% $t$-intervals), 6 normal quarters (circles), and 12 changed quarters (squares) versus the Fixed Buying Rule (+\$6,412), Scripted Expert Buyer (+\$10,795), and Solver (+\$16,923); open weights marked \texttt{*}. (b) Produce bought versus profit relative to the Fixed Rule across the 9-model behavioral cohort (485 runs, separated by model; overall Spearman $\rho = +0.84$). (c) Diagnosis-to-action funnel across 323 quarters with reactive accounts (customer-loss and supplier-shock tasks, 9 models $\times$ 12 tasks) and mid-quarter course-change rates versus 162 normal runs.}
  \label{fig-frontier}
\end{figure*}

Across all 16 frontier models and 859 scored runs, only three models finish ahead of the Fixed Buying Rule (+\$6,412 per quarter) on average (Figure~\ref{fig-frontier}a), led by GPT-6 Astra (+\$7,749), Gemini 4 Carbon (+\$7,703), and Claude Fable 5.2 (+\$7,275), down to GPT-6 Luna at \negusd{8,667}. None beats the Fixed Buying Rule reliably in every run, and every model trails the Scripted Expert Buyer (+\$10,795, 64\% of the Solver's +\$16,923) by more than \$3,000 per quarter.

Models often record a relevant diagnosis without making the corresponding change in later purchases. Across the 485-run behavioral cohort, every model buys 5\% to 42\% less produce by weight than the Fixed Buying Rule, and runs that buy less earn less within every model as well as across models ($\rho = +0.84$, Figure~\ref{fig-frontier}b). Across 136 cases where a model wrote down its own demand forecast, it stocked below 60\% of that estimate in 107 cases (78.7\%). After seven consecutive stockout days on broccoli, GPT-6 Astra recorded the shortage in its notes, continued ordering well below supplier offers, and stayed out of stock on nine of the next ten days.
\begin{tracequote}
Broccoli has stocked out for seven consecutive days, so I'll test a modest increase in its forecast while trimming tomorrow's tomato, grape, carrot and potato orders.
\end{tracequote}
Following product-weeks out of stock at least 25\% of store hours, GPT-6 Astra and GPT-6.1 Sol added +0.09 and +0.07 days of stock, raising orders when their own notes called for restocking in 3 of 26 and 5 of 24 cases. Replaying those episodes with the noted orders placed increased quarterly net value by 1.3\% to 2.7\% of sales, confirming under the simulator that executing the recorded restock plan improved profit.

\begin{figure*}[t]
  \centering
  \includegraphics[width=\textwidth]{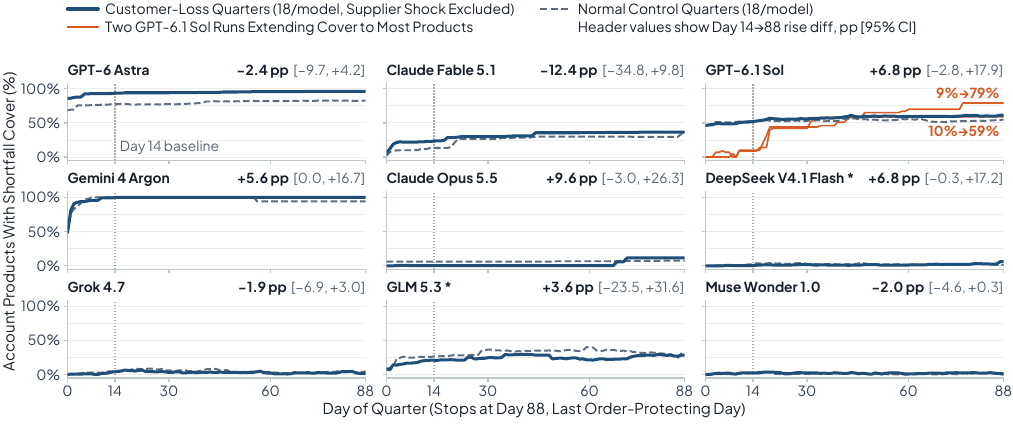}
  \caption{Controlled test of adaptation over time across the 9-model behavioral cohort (Days 0--88, common 0\%--100\% axes). Each panel compares daily share of account products with standing shortfall cover in customer-loss quarters alone (solid; 6 tasks, 18 runs per model, excluding supplier-shock quarters) against the same model's own normal-quarter controls (dashed), reporting the difference in the increase from Day 14 to Day 88 and its 95\% bootstrap interval. Thin lines in GPT-6.1 Sol show the two individual runs extending cover across most products (9\% to 79\% and 10\% to 59\%).}
  \label{fig-adaptation}
\end{figure*}

{\spaceskip=0.33em plus 0.45em minus 0.1em Delayed customer loss shows the same gap between recorded diagnosis and timely action. Across 323 quarters with reactive accounts (customer-loss and supplier-shock tasks), models tied falling orders to their own shortfalls in writing in 69 runs (21.4\%) and protected all six accounts within three days in 12 runs (3.7\%, Figure~\ref{fig-frontier}c). Grok 4.7 recorded that connection in 26 runs and acted within three days in zero while exposed account orders fell past 80\%; Claude Opus 5.5 attributed falling orders to external demand on Day 33 before enabling cover five weeks late on Day 71. Mid-quarter course changes were nine percentage points more frequent in quarters with reactive accounts than in normal quarters (20\% versus 11\%).\par}

Comparing daily shortfall protection in customer-loss quarters against each model's own normal-quarter control separates three distinct patterns (Figure~\ref{fig-adaptation}). First, by Day 14, Gemini 4 Argon has reached full cover and GPT-6 Astra already covers 93.5\% of account products, leaving little room for a further rise during the measured window; this comparison measures later adaptation and does not exclude learning before Day 14. Second, several models drift upward for reasons also present in calm quarters; Claude Fable 5.1 raises cover by +13.2 points in customer-loss quarters versus +25.6 in normal quarters (\negnum{12.4} difference). Third, isolated runs respond directly to customer attrition without producing a dependable model-wide shift. GPT-6.1 Sol rises +6.8 points more from Day 14 in customer-loss quarters than in normal quarters (95\% interval \negnum{2.8} to +17.9), driven by two runs climbing from 9\% to 79\% and 10\% to 59\%, and across all nine models (mean difference +1.5 points; Table~\ref{tab-cover-all} in Appendix~\ref{app-benchmark}) zero models achieve a 95\% interval strictly above zero over the Day 14 to Day 88 window. In the clearest individual case, one Gemini 4 Argon run covered two accounts on Day 2, compared them on Day 10 against four exposed accounts whose orders had halved, and protected the remaining four that morning.
\begin{tracequote}
On Days 1..9 when accounts 1, 4, 5, 6 experienced delivery shortfalls, their daily order volumes permanently dropped by \textasciitilde40-50\%, whereas accounts 2 and 3 (covered since Day 2) maintained 100\% of their order volume!
\end{tracequote}

\section{Learning From the Environment}
\label{sec-learning}

We next study how SMB Bench's records and deterministic replay can improve an open model (Qwen3.8-27B, rank-32 LoRA). Each business group serves a single role. Businesses 1 to 4 supplied operational records for rule design and archived starting states for training, Business 5 supported prompt and optimizer development and selection, and Business 6 served as a sealed test business evaluated once at campaign close.

\begin{figure*}[t]
  \centering
  \includegraphics[width=\textwidth]{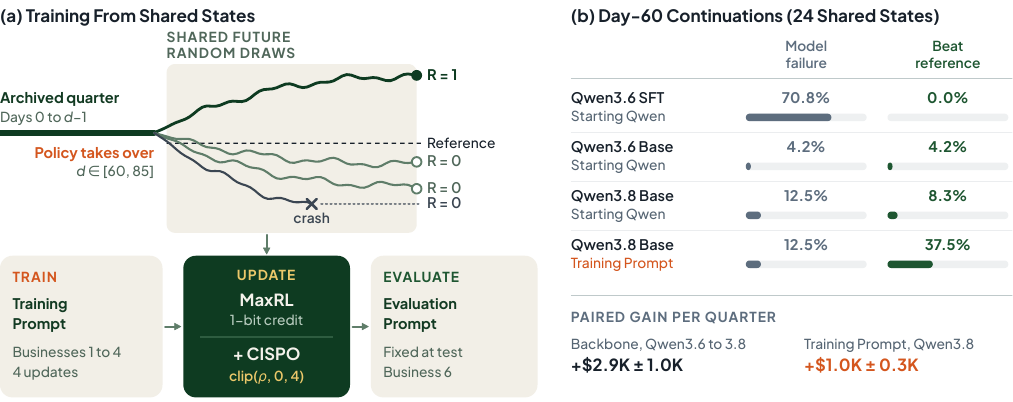}
  \caption{Continuing from archived starting states versus copying transcripts. (a) Replaying Days $0$ to $d-1$ for a morning $d \in [60, 85]$ restores state $s_d$ so four policy continuations share identical history and future random draws with the archived reference run; training runs under the Training Prompt on Businesses 1--4, while evaluation holds the Evaluation Prompt fixed on sealed Business 6. (b) Failure and reference-beat rates across 24 shared Day-60 states, showing 70.8\% failure under action-transcript SFT and paired gains from the Qwen3.8 backbone (+\$2,944 $\pm$ \$1,038) and operational instructions (+\$959 $\pm$ \$332).}
  \label{fig-support}
\end{figure*}

\subsection{Instructions and Training Conditions}
\label{sec-support}

Early baselines clarified how experience must be formatted under delayed consequences (Appendix~\ref{app-negative}). Optimizing single-item replenishment proxies raised daily proxy scores while lowering 90-day profit because minimum-order dispatch fees, multi-day lot ages, and shared cash couple mornings together. Fine-tuning Qwen3.6-35B on 65,235 public tool calls from 44 high-profit frontier runs cut held-out token loss from 0.499 to 0.273, yet terminated early on malformed XML tags, tool-contract errors, or output-token-limit overflows in 17 of 24 (70.8\%) Day-60 continuations and beat the archived reference run in 0 of 24 cases (Figure~\ref{fig-support}b). A plausible explanation is that public archives omit private provider reasoning while keeping a 25-call-per-day action cadence, compounding multi-turn syntax risk when imitated without state-tracking reasoning. Continuing from an archived morning state avoids copying opaque action streams while preserving a consequential state and exact quarter grading.

Under reward-weighted likelihood updates \cite{song2026unified}, policy gradients reweight sampled continuations; when successful continuations are rare, a small rollout group may contain zero successes, leaving no binary-reward contrast for the update. On 24 shared Day-60 states under the Starting Qwen prompt (benchmark rules plus Qwen XML tool syntax), upgrading Qwen3.6 to Qwen3.8 lifted return by +\$2,944 $\pm$ \$1,038 (reference-beat rate 4.2\% to 8.3\%), and adding SQL stockout-adjustment formulas and wholesaler advice (the \emph{Training Prompt}) lifted return by +\$959 $\pm$ \$332 (beat rate 37.5\%, Figure~\ref{fig-support}b).

Full 90-day development on Business 5 then refined those instructions into the \emph{Evaluation Prompt} by adding explicit run-safety limits ($\le 15$ calls per batch, SQL sums) alongside two quantitative rules read from Businesses 1--4 records, standing Day-0 cover across all accounts and capping stock plus arrivals within pre-markdown sales (full validation progression and instruction ablations in Appendix~\ref{app-val}, Table~\ref{tab-prompt-ablations}).

Reinforcement learning ran under the Training Prompt before the run-safety limits and final quantitative rules were introduced (verbatim clauses in Appendix~\ref{app-prompts}). Both untrained and trained Qwen3.8 were then evaluated on sealed Business 6 under the identical Evaluation Prompt; because Starting-Qwen-prompt Qwen3.8 was not evaluated on Business 6, the sealed test measures the final system's standing and the trained-versus-untrained difference under fixed evaluation instructions rather than start-to-final progression.

\subsection{Shared-State Training and Sealed-Test Results}
\label{sec-maxrl-results}

From archived frontier runs on Businesses 1 to 4, we sample a morning $d \in [60, 85]$, deterministically replay Days $0, \dots, d-1$ to restore state $s_d$, and generate $G = 4$ continuations $\tau_1, \dots, \tau_G$ to Day 89 (Figure~\ref{fig-support}a). Because pre-branch history and all subsequent exogenous random draws match across continuations from the same state and the archived reference run $\tau^{\mathrm{ref}}$, shared prefix profit cancels in $G(s_d, \tau_i) - G(s_d, \tau^{\mathrm{ref}})$ \cite{ross2014aggrevate,chang2015lols,kazemnejad2024vineppo}.

Rather than centering raw dollar returns via RLOO \cite{kool2019buy4,ahmadian2024rloo}, we score each continuation with a 1-bit indicator against that morning's reference profit $G_{\mathrm{ref}}(s_d)$ using pass-rate-normalized MaxRL advantages \cite{tajwar2026maxrl},
\begin{align}
R_i &= \mathbb{I}\!\big[\mathrm{valid}(\tau_i) \wedge G(s_d, \tau_i) > G_{\mathrm{ref}}(s_d)\big], \nonumber \\
A^{\mathrm{MaxRL}}_i &= \frac{R_i - \bar{R}}{\bar{R} + \epsilon}, \label{eq-maxrl}
\end{align}
with $\bar{R} = \frac{1}{G}\sum_{j=1}^G R_j$ and $\epsilon = 10^{-6}$. Crashes and sub-reference continuations both score $R_i = 0$; uniform $0/4$ or $4/4$ state groups receive $A^{\mathrm{MaxRL}}_i = 0$ and contribute zero policy-gradient signal; and rare $1/4$ wins receive $+3.0$ versus $-1.0$ for negative continuations. With token importance ratio $\rho_{i,t}(\theta) = \pi_\theta(u_{i,t} \mid h_{i,t}) / \pi_{\mathrm{old}}(u_{i,t} \mid h_{i,t})$, CISPO \cite{minimax2025m1} clips detached weights at $c_{\max} = 4.0$,
\begin{align}
\mathcal{L}_{\mathrm{CISPO}}(\theta) &= -\sum_{i=1}^G \sum_{t \in \mathcal{S}_i} \mathrm{sg}\!\big(\mathrm{clip}(\rho_{i,t}(\theta), 0, c_{\max})\big) \nonumber \\
&\qquad \times A^{\mathrm{MaxRL}}_i \log \pi_\theta(u_{i,t} \mid h_{i,t}). \label{eq-cispo}
\end{align}
Equation~(\ref{eq-cispo}) applies the continuation's advantage to the sampled tokens $\mathcal{S}_i$ across the 5 to 30 buying mornings from $d \in [60, 85]$ to Day 89.

Across 4 updates of 16 continuations (4 states $\times$ 4 rollouts per step), 8 of 16 state groups produced mixed outcomes with non-zero gradients (step curves and rollout diagnostics in Appendix~\ref{app-negative}, Figure~\ref{fig-maxrl-app}). Within the 7 mixed groups containing completed runs on both sides of the reference profit, continuations exceeding the reference profit averaged +\$721 higher return, earning +\$485 more sales while incurring less spoilage (\negusd{155}), lower local cover spend (\negusd{171}), fewer shortfalls (\negnum{13.3~kg}), and smaller Days 85--89 orders (\negnum{100~kg}) than lower-profit continuations from the same state (Table~\ref{tab-siblings}). By contrast, on the identical starting states across updates 0 through 3, continuous-dollar RLOO fell from 100\% to 69\% completion versus 94\% for MaxRL (comparing the complete binary-MaxRL and dollar-RLOO recipes; extended trajectory in Appendix~\ref{app-negative}, Figure~\ref{fig-maxrl-app}b). Because episodes terminating on parse or token errors receive a world's \negusd{65,000} failure floor, a single crashed continuation gave surviving RLOO continuations from the same state a $+15.32$ mean advantage even when they trailed the reference profit by >\$3,000, offering a plausible explanation for why verbose near-timeout outputs multiplied under dollar centering while 1-bit reference comparisons kept positive-advantage outputs shorter than negative ones (4,868 versus 7,065 tokens per call).

\begin{table}[t]
  \caption{Four continuations from the same Day-60 supplier-shock state on Business 2 (full cover, 0~kg prior shorts, Gemini 4 Argon archived reference profit \negusd{3,632}). Post-branch local spend and shortfalls cover Days 60--89.}
  \label{tab-siblings}
  \vskip 0.04in
  \centering
  \resizebox{\columnwidth}{!}{%
  \begin{tabular}{lcccc}
    \toprule
    Continuation & d60--62 POs & Post Short & Local Spend & Quarter Profit \\
    \midrule
    Continuation 1 ($A = +1$) & 88 (1,364 kg) & 19.4 kg & \$1,997 & \negusd{2,725} (+\$907) \\
    Continuation 2 ($A = +1$) & 66 (956 kg) & 29.5 kg & \$2,893 & \negusd{2,883} (+\$748) \\
    Cover reversal ($A = -1$) & 43 (667 kg) & 109.7 kg & \$2,162 & \negusd{3,725} (\negusd{94}) \\
    Cash exhaustion ($A = -1$) & 41 (642 kg) & 60.0 kg & \$3,960 & \negusd{3,737} (\negusd{105}) \\
    \bottomrule
  \end{tabular}%
  }
\end{table}

\begin{figure*}[t]
  \centering
  \includegraphics[width=\textwidth]{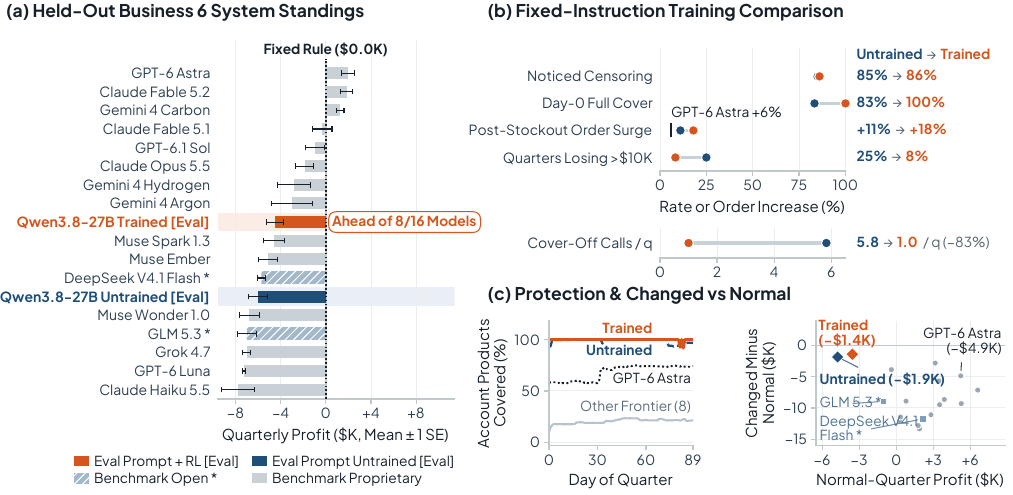}
  \caption{Held-out Business 6 system placement and fixed-instruction training comparison. (a) Business 6 profit ($\pm$ 1 SE) for Qwen3.8-27B systems under the Evaluation Prompt (\texttt{[Eval]}) beside 16 leaderboard models under the Benchmark Prompt (open weights \texttt{*}). (b) Holding the unseen Evaluation Prompt fixed, the trained policy keeps written stockout recognition unchanged (85\% vs.\ 86\%) while lifting Day-0 cover (83\% to 100\%) and post-stockout order responses (+11\% to +18\%) and reducing protection reversals (5.8 to 1.0 per quarter) and >\$10K loss quarters (25\% to 8\%). (c) Daily account cover and profit change relative to normal quarters (\texttt{Other Frontier (8)} averages the remaining eight behavioral-cohort models).}
  \label{fig-test}
\end{figure*}

{\spaceskip=0.33em plus 0.45em minus 0.1em On sealed Business 6, where both policies completed all 12 quarters without failure penalties (Figure~\ref{fig-test}a; per-quarter outcomes in Table~\ref{tab-individual-runs}), the complete Qwen3.8-27B system remains loss-making overall (\negusd{4,495} $\pm$ \$767 under the Evaluation Prompt) yet finishes ahead of 8 of 16 leaderboard entries running the Benchmark Prompt, including DeepSeek V4.1 Flash (\negusd{5,707}), GLM 5.3 (\negusd{7,002}), and Grok 4.7 (\negusd{7,013}), while trailing GPT-6 Astra (+\$1,969). Holding the unseen Evaluation Prompt fixed against untrained Qwen3.8 (\negusd{5,984} $\pm$ \$834), the estimated profit difference was +\$1,489 $\pm$ \$1,133 ($\mathrm{SE}_{\Delta} = \sqrt{\mathrm{SE}_{\mathrm{trained}}^2 + \mathrm{SE}_{\mathrm{untrained}}^2}$). While measured recognition of stockout effects in \texttt{\textless{}think\textgreater{}} changed little (84.7\% to 85.6\%, Figure~\ref{fig-test}b), the trained policy followed customer-protection commitments more consistently across the quarter. Next-day orders after $\ge$4h stockouts rose to 1.18$\times$ item mean (versus 1.11$\times$ untrained and 1.06$\times$ Astra), Day-0 full cover on Business 6 rose from 83\% (10/12) to 100\% (12/12) of quarters, mid-quarter cover-off reversals fell from 5.8 to 1.0 per quarter (\negnum{83\%}), and quarters losing over \$10,000 fell from 25\% (3/12) to 8\% (1/12). In the untrained model's worst test quarter (\negusd{13,913}), a single normal onion order after a partial short prompted an incorrect explanation followed by 31 cover-off calls across Days 73--74, causing 441~kg of shorts and \negusd{4,177} in lost-customer reserve.\par}
\begin{tracequote}
a4 onions: f ≈ 0 → MV = penalty 295 - premium 3,305 = -3,010 → should OFF?!?! This is a 180-degree reversal from earlier analysis!
\end{tracequote}
Under the identical unseen instructions, the trained policy issued zero cover-off calls across all normal and customer-loss test quarters and lost \negusd{1,408} between changed and normal conditions (Figure~\ref{fig-test}c).

\section{Remaining Coordination Failure}
\label{sec-coordination}

\begin{figure*}[t]
  \centering
  \includegraphics[width=\textwidth]{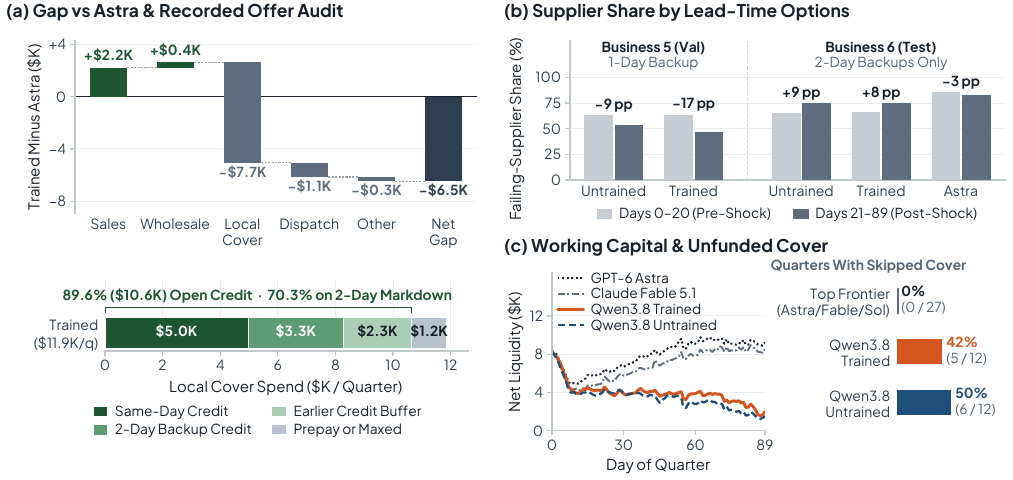}
  \caption{Upstream supply, lead-time, and working-capital bottlenecks on Business 6. (a) Profit gap versus GPT-6 Astra (\negusd{6,465}) and audit of recorded supplier offers and purchasing constraints across the trained policy's \$11,868/quarter emergency local spend, showing 89.6\% (\$10,630) occurred while open funded credit offers existed (70.3\% on 2-day markdown items). (b) Failing-supplier order share pre- versus post-Day 21 on Business 5 (\negnum{17pp}, 1-day backup available) versus Business 6 (+8pp, 2-day backups only). (c) Daily net liquidity and unfunded (\texttt{skipped\_funds}) cover failures.}
  \label{fig-coord}
\end{figure*}

Protecting customers consistently exposed an upstream requirement across ordering, supplier lead times, and cash. On Business 6, the trained policy received supply equivalent to 88.8\% of total demand, close to GPT-6 Astra's 88.6\%, yet fulfilled less walk-in demand (74.1\% versus 80.3\%), spent \$11,868 versus \$4,190 per quarter on 6 a.m.\ emergency local purchases across 72 of 90 days, and trailed Astra by \$6,465 (Figure~\ref{fig-coord}a, top). Checking recorded supplier offers and morning purchasing constraints (\texttt{available\_\allowbreak to\_\allowbreak commit}) across all 12 test quarters shows that cooler limits (0.0\%), case caps (0.7\%), and 3-day prepay terms (9.7\%) rarely caused these local buys (Figure~\ref{fig-coord}a, bottom). Instead, 89.6\% (\$10,630 per quarter) occurred while open, funded credit (\texttt{NET}) wholesale offers sat unbought within physical shelf life, split across same-day credit (41.9\%), 2-day backup credit after truck slips (27.9\%), and earlier buffer lots (19.8\%).

Order logs and reasoning traces point to a plausible explanation in how the prompt's pre-markdown lot cap interacted with available suppliers and their lead times. Twelve of 20 items on Business 6 reduce walk-in shelf price after two days (\texttt{markdown\_after\_days = 2}) despite 7 to 28 days of physical shelf life and fixed account contract prices (spoilage was \$488, with 0.70\% of received weight expired or culled); 70.3\% (\$8,342) of emergency spend fell on those 12 items, and store-wide inventory averaged 0.89 days against a 1.32-day supplier lead time.
\begin{tracequote}
asp 1 is enough; let the cover handle asp line gaps (accept the premium, protect the account)
\end{tracequote}
Holding 0.89 days of stock is also consistent with difficulty switching to slower backup suppliers on Business 6, where supplier switching stalled after succeeding on Business 5 (Figure~\ref{fig-coord}b). Business 5 offered an alternative 1-day credit vendor, and the trained buyer reduced the deteriorating supplier's share by 17 points after Day 21 (63.8\% to 46.6\%). On Business 6, the deteriorating vendor (capped at 280.9~kg/day) was the sole 1-day supplier while reliable credit backups required 2 days; although notebooks recorded its 15\%--20\% cancellation rate, its order share still rose from 66.8\% to 74.6\% (whereas Claude Fable 5.1 held 1.40 days of stock across 2-day suppliers with \$10 local spend).

Repeated 2.54$\times$ cash local purchases drained net liquidity to a +\$1,177 minimum versus Astra's +\$4,659 (Figure~\ref{fig-coord}c), causing local cover to fail unfunded (\texttt{skipped\_funds}) in 42\% (5/12) of trained and 50\% (6/12) of untrained quarters versus 0\% (0/27) for top frontier models, and triggering a Day-83 cover pause in the trained policy's single >\$10,000 loss quarter (\negusd{12,657}). Keeping standing local cover active throughout calm normal quarters also lowered normal-quarter return (\negusd{3,557} versus Astra's +\$5,218). Adequate aggregate supply and consistent customer protection still fell short without multi-day wholesale buffers and cash discipline to support that commitment.

\section{Related Work}
\label{sec-related}

\subsection{Stateful and Economic Agent Benchmarks}
Interactive coding, web, and tool benchmarks test multi-step execution across bounded tasks \cite{jimenez2024swebench,zhou2024webarena,yao2024taubench,trivedi2024appworld,xie2024osworld,kwa2025measuring}, while recent stateful benchmarks examine sustained commercial operation and adaptation \cite{backlund2025vending,zheng2025lifelong}. StoreBench combines a persistent commerce environment, replay, operational diagnostics, and post-training \cite{raghuvanshi2026storebench}; CEO-Bench studies delayed business effects across coordinated decisions \cite{chen2026ceobench}; and Continual Learning Bench compares stateful against stateless runs to separate accumulated experience from starting competence \cite{asawa2026clbench}. Close behavioral studies likewise examine where diagnosis and action diverge, as STOCKTAKE compares stated diagnosis against subsequent control \cite{deb2026stocktake} and RetailBench tracks evidence acquisition, action conversion, and follow-up \cite{zhang2026retailbench} alongside multi-stage inference audits \cite{li2026genebench}. Within this agenda \cite{minsky1961steps,silver2025experience}, SMB Bench combines matched Day-0 consequence conditions, missed-action replays, and archived frontier mornings in one study where the buyer's own decisions alter later sales evidence and cash.

\subsection{Interactive Imitation and State-Continuation RL}
Off-policy imitation compounds errors under state drift \cite{ross2011dagger}, whereas branching from visited states under cost-to-go evaluation lets learners improve upon imperfect reference policies (AggreVaTe, \citealp{ross2014aggrevate}; LOLS, \citealp{chang2015lols}; Go-Explore, \citealp{ecoffet2021goexplore}; VinePPO, \citealp{kazemnejad2024vineppo}). Recent agent-training methods build directly on interaction states and restored prefixes, using action-generated future states as supervision \cite{zhang2025earlyexperience}, reconstructing intermediate states from successful trajectories to sample progressive suffix continuations (CRPS, \citealp{zhang2026crps}), checkpointing environment and context snapshots for verifier-guided branching (SPROUT, \citealp{liu2026sprout}), and forking rollouts for process credit or distillation (Counterfactual Rollout Replay, \citealp{li2026crr}; \citealp{li2026daggerllm,luo2026rollback,zhang2026stateopd}). Building on restored-state continuation, our study branches from archived frontier mornings against reference economic profit and tests which operational behaviors transfer to an unseen business under fixed instructions.

\subsection{Prompt Support and Likelihood-Grounded RL}
In-context reflection and cross-episode learning methods extract reusable guidance or adapt online across episodes \cite{shinn2023reflexion,wang2023voyager,lin2026orbit,shalev2026reflector}, while \citet{song2026unified} analyze post-training through reward-weighted likelihood over sampled rollouts. Within critic-free group objectives \cite{kool2019buy4,ahmadian2024rloo}, Dr.\ GRPO \cite{liu2025drgrpo} and DAPO \cite{yu2025dapo} address length biases on single-turn reasoning tasks, whereas binary reference-profit rewards (MaxRL, \citealp{tajwar2026maxrl}) remove dependence on episode-failure penalty magnitudes and CISPO \cite{minimax2025m1} clips the importance weights used in the policy update.

\subsection{Operations Research and Supply-Chain Agents}
Inventory theory characterizes censored demand estimation \cite{huh2009censored,besbes2013censoring,wang2025freshretailnet}, perishable lot lifetimes \cite{nahmias1982perishable,nahmias2011perishable}, and dual sourcing between costly emergency supply and longer-lead wholesale orders \cite{whittemore1977optimal,veeraraghavan2008now,xin2018tailored}. Relative to stylized inventory gyms and negotiation testbeds \cite{hubbs2020orgym,liu2026sabreagent,long2026supplychain}, SMB Bench couples SQL observability, customer retention, supplier reliability, and working capital in a single replayable buying desk.

\section{Discussion and Limitations}
\label{sec-discussion}

SMB Bench uses the same environment to measure failures to act on delayed consequences, test the consequences of missed actions through replay, and improve the resulting decisions. Behavioral audits and replays distinguish recording a diagnosis, consistently acting on it, and improving business profit. We used the environment's records to identify operational rules and its archived states to create training experience; on an unseen business under identical evaluation instructions, the trained policy followed customer-protection commitments more consistently with fewer severe-loss quarters, while the additional profit benefit remained uncertain. Persistent purchasing and cash-management failures showed what the agent still needed to coordinate to make those commitments economical.

\subsection{Benchmark Limitations}
Three boundaries govern interpretation of the benchmark results. First, all six businesses instantiate a single simulated perishable-wholesale workflow with specified customer-attrition and supplier-degradation rules rather than live firm deployments. Second, our 42-to-18 screening protocol deliberately selects delayed mechanisms with measurable profit impact ($\ge 0.25\%$ of baseline revenue in at least four stores) across six synthetic businesses rather than an unweighted sample of routine store events. Third, buyer notebooks and provider-returned reasoning summaries are incomplete records of the causes behind model actions (Appendix~\ref{app-negative}).

\subsection{Learning Study Limitations and Future Work}
Three qualifications bound the learning study. First, validation (Business 5) and test (Business 6) comprise one business each, where 12 test quarters resolve roughly \$2,300 at 2 SE; although all four fixed-instruction comparisons across those two businesses had positive point estimates using the same trained checkpoint, the sealed-test profit difference (+\$1,489 $\pm$ \$1,133) should be read alongside reduced protection reversals and large-loss quarters rather than as a standalone $p < 0.05$ single-store margin. Second, leaderboard models used the Benchmark Prompt while our complete Qwen3.8 system used the Evaluation Prompt (not run at initial-prompt baseline on Business 6), so leaderboard rank compares complete systems whereas fixed-instruction pairs isolate the weight update; early SFT and RL comparisons also differed in model size (Qwen3.6-35B for SFT) and training instructions (Training Prompt for RL; details in Appendices~\ref{app-negative} and \ref{app-repro}). Third, researchers extracted the quantitative rules from Businesses 1--4 records and our recorded-offer audit checks individual local buys without dynamic order replay. Extending trained-reflector and cross-episode learning methods \cite{lin2026orbit,shalev2026reflector} alongside earlier morning continuations (Days 12--59) raises a concrete next question, whether an agent can autonomously extract candidate operational rules, test their downstream supply and cash costs, and revise its behavior without human interpretation.

\section*{Acknowledgments}
We thank GetASAP and participating wholesale produce operators for workflow consultations and domain checks during the design of SMB Bench.

{\let\oldthebibliography\thebibliography
\renewcommand{\thebibliography}[1]{\oldthebibliography{#1}\setlength{\parskip}{0pt}\setlength{\itemsep}{3.5pt plus 0.8pt minus 0.8pt}\spaceskip=0.33em plus 0.38em minus 0.1em\relax}
\bibliography{references}
\bibliographystyle{icml2026}
}

\newpage
\appendix
\renewcommand{\thefigure}{A\arabic{figure}}
\setcounter{figure}{0}
\onecolumn
\raggedbottom
\setlength{\parskip}{6pt}
\section{Tacit Knowledge, Negative Results, and Optimizer Diagnostics}
\label{app-negative}

Following \citet{polanyi1966tacit}, we document the failed post-training shortcuts, prompt regressions, continuous-dollar RLOO crash-penalty dynamics, and unexpected frontier agent behaviors encountered across the 859-run benchmark evaluation and the 94-experiment training and evaluation program.

\subsection{Single-Decision Proxy Reward Goodharting}
Before training full-quarter continuations, our initial post-training campaign isolated a single operational sub-skill, stockout-aware onion replenishment, scored at individual morning decision points against local supported-action margins relative to the Fixed Buying Rule. Optimization reliably increased single-decision proxy rewards on the training business, yet full 90-day net business value declined and gains failed to transfer to held-out stores. Wholesaler workflow audits (\texttt{GetASAP} consultation records) and mass-balance checks explained why. Supplier minimum-order dispatch thresholds couple products ordered on the same truck, multi-day lot ages couple consecutive mornings' spoilage and markdown exposure, and shared working capital couples wholesale orders to automatic local cover. Local decision proxies rewarded aggressive single-item ordering that triggered storage dock refusals (up to 306 refused cases per quarter in early supervised prototypes) without improving whole-desk profit. Grading complete continuations to Day 89 closes this gap by construction, because refusal fees, dispatch fees, and lot write-offs enter the same quarter profit (Equation~\eqref{eq-profit}) that defines the reward.

\subsection{Reasoning-Stripped Frontier SFT Collapse}
To test off-policy imitation of strong buyers, we constructed a supervised fine-tuning dataset of 65,235 public tool-call targets from the 44 highest-profit training-store frontier runs (Businesses 1 to 4) and fine-tuned Qwen3.6-35B-A3B. Held-out token negative log-likelihood dropped cleanly from 0.499 to 0.273. Yet when evaluated from 24 shared Day-60 starting states under temperature-1 sampling and strict tool parsing, 17 of 24 runs (70.8\%) terminated early on model errors (11 XML tool-parse failures such as unclosed parameter tags, 4 tool-contract violations such as bundling \texttt{advance\_day} with other calls, and 2 output-token-limit overflows), exceeding the archived reference profit in 0 of 24 continuations. Rollout forensics identified a mismatch between public action traces and reasoning-model execution. Public benchmark archives strip private provider chain-of-thought (replacing hidden reasoning with empty \texttt{\textless{}think\textgreater{}\textless{}/think\textgreater{}} blocks) while preserving frontier tool cadence (GPT-6 Astra averages 25.6 model calls per day versus Qwen3.8's 7.4). Copying unbatched 40-to-60-call-per-day action streams without the internal reasoning that tracked state across turns offers a plausible explanation for why per-call syntax hazard compounded across 830 to 1,319 calls over 11 to 28 days. On the 7 of 24 continuations that completed without syntax errors, the SFT checkpoint outperformed matched Qwen3.6 Base by +\$1,284 $\pm$ \$1,763.

\subsection{Continuous-Dollar RLOO Survivor Skew Versus Binary MaxRL}
\label{app-rloo}

\begin{figure}[!htbp]
  \centering
  \includegraphics[width=\textwidth]{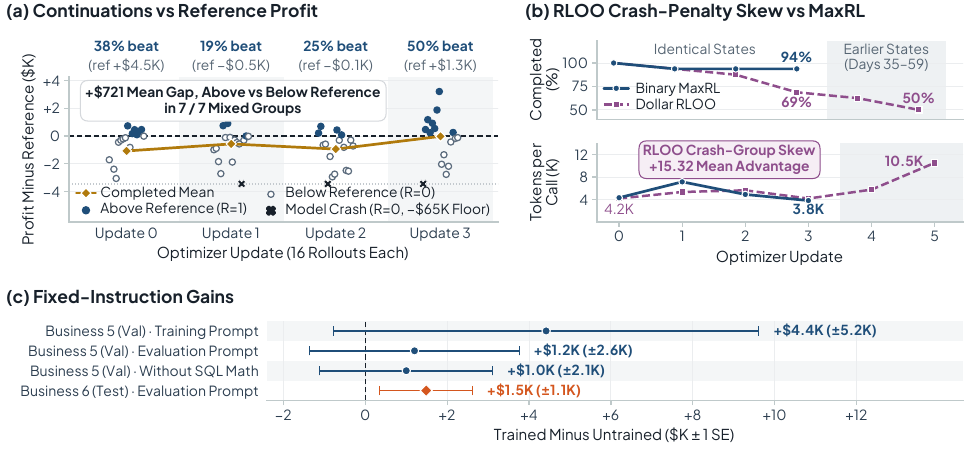}
  \caption{Optimizer diagnostics comparing binary MaxRL against continuous-dollar RLOO on identical starting states. (a) All 64 MaxRL continuations across 4 updates relative to each morning's archived reference profit, marking reference profit tiers, completed-run means (\texttt{Completed Mean}), and the +\$721 gap between continuations above and below reference across all 7 mixed completed groups. (b) On identical states in updates 0--3, binary MaxRL maintains 94\%--100\% completion while dollar RLOO falls to 69\%; after two further RLOO updates from earlier states (Days 35--59), RLOO reaches 50\% completion and 10,506 tokens/call by update 5 under $+15.32$ crash-group survivor skew. (c) Fixed-instruction gains across all four held-out comparisons involving one trained checkpoint and two businesses.}
  \label{fig-maxrl-app}
\end{figure}

Figure~\ref{fig-maxrl-app} compares 1-bit reference-profit MaxRL against continuous-return RLOO \cite{kool2019buy4,ahmadian2024rloo}, $A^{\mathrm{RLOO}}_i = \frac{G}{G-1}(G_i - \bar{G})$, on the exact same 16 starting states in updates 0 to 3. The two recipes used identical starting states for updates 0 through 3; the dollar-reward RLOO run continued for two additional updates from earlier states (Days 35--59), reaching 50\% completion and 10,506 tokens per call by update 5. The comparison therefore evaluates the complete binary-MaxRL and dollar-RLOO recipes rather than isolating reward binarization from advantage normalization. Returns $G$ in Equation~\eqref{eq-rloo-skew} are scaled in thousands of dollars (\$K), matching the training reward. In multi-turn environments where a malformed tool call or 65,536-token overflow terminates the episode at a world's failure floor $G_{\mathrm{crash}} \approx -65$ (\negusd{65}K), a single crashed continuation while $G - 1 = 3$ continuations finish at mediocre returns $G_{\mathrm{med}}$ well below the reference profit gives every surviving continuation a positive advantage,
\begin{equation}
\label{eq-rloo-skew}
A^{\mathrm{RLOO}}_{i > 1} = \frac{G_{\mathrm{med}} - G_{\mathrm{crash}}}{G - 1} > 0.
\end{equation}
At $G = 4$ and $G_{\mathrm{med}} - G_{\mathrm{crash}} \approx 45$ (\$45K), Equation~\eqref{eq-rloo-skew} assigns surviving continuations roughly $+15$ advantage (+15.32 empirical mean across 26 crash-group survivors, including 7,108-token/call continuations trailing the reference profit by more than \$3,000). Positive advantages on verbose survivors coincided with mean output length rising from 4,181 to 10,506 tokens per call and completion falling from 100\% to 50\% by update 5 (Figure~\ref{fig-maxrl-app}b), although updates 4 and 5 also drew earlier starting states. Across the 4 MaxRL updates (16 state groups, Figure~\ref{fig-maxrl-app}a), 7 groups went $0/4$ against strong reference runs, 1 went $4/4$ against a weak reference run, and 8 produced mixed outcomes (pass rates of 38\%, 19\%, 25\%, and 50\% across updates 0--3).

\subsection{Validation Development Progression and Instruction Regressions}
\label{app-val}
Table~\ref{tab-prompt-ablations} reports all nine settings evaluated during development on Business 5 (validation), where moving from Starting-Qwen-prompt Qwen3.8 (\negusd{8,599} $\pm$ \$1,904, 78\% completion) to the Evaluation Prompt plus shared-state RL (+\$2,419 $\pm$ \$760, 100\% completion) raised validation return by +\$11,018 per quarter (partly reflecting reduced early terminations scored at \negusd{25,000}). Three non-additive regressions emerged during prompt design. First, adding uncapped wholesaler advice (the Training Prompt) on top of SQL censoring formulas dropped completion from 78\% to 58\% (\negusd{15,188} $\pm$ \$3,388), as four of the five failed quarters overflowed the 65,536-token per-call ceiling and one hit repeated tool-contract errors while spoilage doubled from \$6.3K to \$12.8K; explicit run-safety limits ($\le 15$ tool calls per batch, SQL sums) restored 100\% completion (\negusd{967} $\pm$ \$1,315), and replacing conditional cover with standing Day-0 cover and the pre-markdown lot cap lifted untrained return to +\$1,216 $\pm$ \$2,447 (swinging customer-loss quarters by +\$10,616 while costing \negusd{9,346} in normal quarters). Second, removing SQL demand-censoring formulas starved wholesale volume and pushed calm-quarter local spending to \$29.5K (\negusd{5,087} overall). Third, restoring the conditional \emph{unless clearly uneconomic} clause revived item-by-item cover toggling, re-introducing 456~kg of shortfalls (\negusd{4,473}). Holding instructions fixed on Business 5, the trained checkpoint exceeded untrained Qwen3.8 by +\$4,418 $\pm$ \$5,198 on the Training Prompt, +\$1,203 $\pm$ \$2,562 on the Evaluation Prompt, and +\$1,002 $\pm$ \$2,114 without SQL formulas (Figure~\ref{fig-maxrl-app}c, with $\mathrm{SE}_{\Delta} = \sqrt{\mathrm{SE}_{\mathrm{trained}}^2 + \mathrm{SE}_{\mathrm{untrained}}^2}$). Under the Evaluation Prompt, that +\$1,203 validation advantage depends on scoring the untrained policy's single Day-75 failed supplier-shock quarter at \negusd{25,000} (11/12 versus 12/12 completed); among completed quarters with equal weight across conditions, the untrained policy averaged +\$3,279 versus +\$2,419 for the trained policy (\negusd{859} $\pm$ \$1,756, Table~\ref{tab-individual-runs}). On the all-run validation score, the policies were within one standard error. The declared rule selected the untrained policy, and both were evaluated on the sealed test business.

\begin{table}[!htbp]
  \caption{Complete Business 5 validation comparison across backbone, the Starting Qwen prompt, the Training Prompt, run-safety limits, the Evaluation Prompt, and continuation RL (early model-failure quarters scored at \negusd{25,000}).}
  \label{tab-prompt-ablations}
  \vskip 0.06in
  \centering
  \resizebox{\textwidth}{!}{%
  \begin{tabular}{llcccccc}
    \toprule
    Validation Arm & Instruction Setting & Completion & Mean (\$/q) & Normal & Customer Loss & Supplier Shock & Shortfall \\
    \midrule
    Qwen3.6 Starting ($n=3$) & Starting Qwen prompt & 1.00 & \negusd{28,963} & \negusd{12,784} & \negusd{51,263} & \negusd{22,843} & 1,005 kg \\
    Qwen3.8 Starting ($n=9$) & Starting Qwen prompt & 0.78 & \negusd{8,599} & +\$15,778 & \negusd{15,743} & \negusd{25,831} & 772 kg \\
    Training Prompt ($n=12$) & + SQL OOS math + wholesaler advice & 0.58 & \negusd{15,188} & \negusd{4,596} & \negusd{15,368} & \negusd{25,600} & 586 kg \\
    + Run-Safety Limits ($n=6$) & + $\le 15$ batch limit + SQL sums & 1.00 & \negusd{967} & +\$13,193 & \negusd{4,377} & \negusd{11,718} & 513 kg \\
    Evaluation Prompt ($n=12$) & + Day-0 cover + pre-markdown cap & 0.92 & +\$1,216 & +\$3,847 & +\$6,239 & \negusd{6,437} & 8 kg \\
    Evaluation Prompt + RL ($n=12$) & + 4 updates shared-state MaxRL & 1.00 & +\$2,419 & +\$4,734 & +\$3,970 & \negusd{1,445} & 16 kg \\
    \midrule
    Without SQL Math ($n=12$) & Safety limits + 2 rules only & 1.00 & \negusd{5,087} & \negusd{2,794} & \negusd{4,218} & \negusd{8,248} & 0 kg \\
    + Supplier SQL Rule ($n=12$) & Above + supplier-switch SQL & 1.00 & \negusd{4,331} & \negusd{3,607} & \negusd{974} & \negusd{8,413} & 0 kg \\
    Conditional Cover ($n=12$) & Evaluation Prompt with cost exception & 1.00 & \negusd{4,473} & +\$2,924 & \negusd{2,630} & \negusd{13,713} & 456 kg \\
    \bottomrule
  \end{tabular}%
  }
\end{table}

\subsection{Unexpected Frontier Behaviors and Reasoning-Summary Omissions}
Auditing all 859 frontier runs and 208,012 reasoning traces surfaced three operational failure modes rarely seen in short-horizon evaluations.

\subsubsection{Silent Memory Loss From Refused Over-Limit Writes}
Memory writes exceeding 4,000 bytes per call return an explicit error instructing the buyer to write the first chunk and append the rest. Gemini 4 Argon tended to rewrite its entire notebook in one call and recovered on the same day from only 44\% of its 1,949 refused writes (versus 91\% to 100\% across all other models). In 15 of its 54 runs, Gemini 4 Argon operated for 40 or more consecutive days without saving a single note, scoring far below its own task mean.

\subsubsection{Identical Exogenous Verdicts Across Causal Twins}
Across paired normal and customer-loss runs of the same store, Grok 4.7 reached the exact same written conclusion that shortfalls do not harm customer demand.
\begin{tracequote}
No evidence from the eight-day history shows that shortfalls reduce future orders, so local cover will remain off until demand destruction is observed.
\end{tracequote}
In the stationary control that belief was harmless; in the customer-loss twin its exposed accounts fell to 11\%--20\% of unshorted order volume while Day-89 notes still listed customer sensitivity as unverified. Similarly, Claude Opus 5.5 dismissed shrinking orders on Day 33 before recording {\spaceskip=0.42em plus 0.25em minus 0.12em\texttt{account demand collapses after shortfalls}} on Day 71.

\subsubsection{Provider Reasoning Summaries Hide Operational Causes}
Most proprietary APIs return model-generated reasoning summaries rather than raw reasoning tokens. Across 208,012 traces, 0.35\% of summaries hallucinated off-task events (1 in 32 for Muse, 1 in 150 for GPT-6.1 Sol and GPT-6 Astra, such as claiming \emph{generated some disallowed content} after a routine stray-argument tool error). On 91 stocking decisions where Sol's and Astra's own notebooks recorded why they held back orders, the provider summaries explained the holdback in 0 of 91 cases, whereas DeepSeek's unsummarized reasoning explained 43 of 45.

\section{Benchmark Construction Details and Full 16-Model Tables}
\label{app-benchmark}

\subsection{Daily Execution Sequence and World Audits}
Within each 90-day episode, overnight and daytime events execute in a fixed deterministic order. At 5 a.m., the buyer receives the daily status dashboard and executes SQL queries, quote inspection, purchase-order placement, shortfall-cover updates, and memory edits until calling \texttt{advance\_day}. Before store open, arriving supplier trucks undergo dock inspection and cold/cool/warm room capacity checks; accepted cases enter lot-tracked inventory while late or pre-cancelled cases log status updates and automatic vendor claims where applicable. At 6 a.m., due account order lines fill oldest-lot-first from on-hand stock; any remaining shortfall either triggers automatic local-market purchase (if covered and \texttt{available\_to\_commit} suffices, otherwise logging \texttt{skipped\_funds}) or ships short with a 5\% fee and updates hidden account/line retention state in reactive worlds. Across the 16-hour store day, walk-in shoppers draw hourly demand against shelf stock, logging \texttt{grams\_sold} and \texttt{hours\_out\_of\_stock}. At evening close, lots crossing \texttt{markdown\_after\_days} step down walk-in retail price (leaving contracted account prices untouched), expired lots dispose to waste, daily overhead and interest accrue, and 21-day or 7-day supplier invoices settle.

Table~\ref{tab-admission} summarizes the 42-candidate screening across 168 audited worlds. Two pre-release scoring checks documented in the public benchmark report ensure clean evaluation. First, the mid-quarter course-change check excludes cover rules already set during Day 0--1 initial setup, reclassifying two Claude Opus 5.5 episodes without altering task admissions. Second, buyer-defined DuckDB SQL column aliases (such as \texttt{AS life\_days}) are exempted from the private-column name filter after auditing 23 flags across 18 runs and confirming zero private table access. Table~\ref{tab-full-leaderboard} reports complete 16-model benchmark performance across dollar profit and revenue-normalized margin points over the Fixed Buying Rule, and Table~\ref{tab-cover-all} reports controlled Day-14-to-Day-88 cover adaptation across all 15 models with exported cover trajectories.

\begin{table}[h]
  \caption{Screening 42 candidate tasks across 168 audited worlds. Admitted consequence tasks cause $\ge 0.25\%$ baseline-revenue damage in $\ge 4/6$ stores, pass controllability, and score $0/0/0$ on shortcut checks.}
  \label{tab-admission}
  \vskip 0.05in
  \centering
  \small
  \begin{tabular}{lcccl}
    \toprule
    Candidate Mechanism & Median Damage & Stores $\ge 0.25\%$ & Shortcuts & Status \\
    \midrule
    Supplier shock (d21) & 3.22\% rev & 6 / 6 & 0 / 0 / 0 & Admitted (6) \\
    Reactive accounts & 1.90\% rev & 6 / 6 & 0 / 0 / 0 & Admitted (6) \\
    Stationary normal & 0.00\% rev & Control & 0 / 0 / 0 & Admitted (6) \\
    \midrule
    Outside demand drop & 0.24\% rev & 2 / 6 & 0 / 0 / 0 & Rejected \\
    Lot shelf-life drop & 0.20\% rev & 1 / 6 & 0 / 0 / 0 & Rejected \\
    Supplier quality drop & 0.16\% rev & 2 / 6 & 0 / 0 / 0 & Rejected \\
    Walk-in reaction & 0.06\% rev & 0 / 6 & 0 / 0 / 0 & Rejected \\
    \bottomrule
  \end{tabular}
\end{table}

\begin{table}[!p]
  \caption{Complete 16-model SMB Bench standings across 859 scored runs on 18 tasks, showing quarterly dollar returns alongside revenue-normalized margin points over the Fixed Buying Rule on the 12 changed-quarter tasks (95\% $t$-intervals over tasks, $\mathrm{df}=11$) and list-price inference cost per run.}
  \label{tab-full-leaderboard}
  \vskip 0.04in
  \centering
  \resizebox{\textwidth}{!}{%
  \begin{tabular}{lccccccc}
    \toprule
    Model & All 18 Tasks & Normal (6) & Changed (12) & vs Fixed Rule & Changed Margin (95\% CI) & Valid Runs & Cost / Run \\
    \midrule
    GPT-6 Astra & +\$7,749 & +\$10,145 & +\$6,551 & +\$1,337 & +2.18 (+0.04 to +4.32) & 54 / 54 & \$213 \\
    Gemini 4 Carbon & +\$7,703 & +\$9,346 & +\$6,881 & +\$1,291 & +2.37 (+0.46 to +4.29) & 53 / 54 & \$134 \\
    Claude Fable 5.2 & +\$7,275 & +\$12,545 & +\$4,640 & +\$863 & +1.00 (\negnum{0.56} to +2.56) & 54 / 54 & \$94 \\
    \emph{Fixed Buying Rule} & \emph{+\$6,412} & \emph{+\$10,051} & \emph{+\$4,592} & \emph{+\$0} & \emph{+0.00 (Reference)} & \emph{18 / 18} & \emph{\$0} \\
    Claude Fable 5.1 & +\$5,427 & +\$9,764 & +\$3,259 & \negusd{985} & +0.16 (\negnum{1.42} to +1.74) & 54 / 54 & \$106 \\
    GPT-6.1 Sol & +\$5,404 & +\$10,331 & +\$2,941 & \negusd{1,008} & \negnum{0.05} (\negnum{2.75} to +2.64) & 54 / 54 & \$32 \\
    Gemini 4 Hydrogen & +\$2,560 & +\$9,336 & \negusd{827} & \negusd{3,851} & \negnum{2.03} (\negnum{4.34} to +0.29) & 52 / 54 & \$25 \\
    Gemini 4 Argon & +\$1,501 & +\$4,597 & \negusd{46} & \negusd{4,911} & \negnum{1.65} (\negnum{3.46} to +0.16) & 54 / 54 & \$70 \\
    Claude Opus 5.5 & +\$907 & +\$9,401 & \negusd{3,340} & \negusd{5,505} & \negnum{3.99} (\negnum{5.81} to \negnum{2.17}) & 54 / 54 & \$23 \\
    DeepSeek V4.1 Flash* & \negusd{3,036} & +\$7,745 & \negusd{8,426} & \negusd{9,448} & \negnum{6.73} (\negnum{8.36} to \negnum{5.09}) & 54 / 54 & \$34 \\
    Grok 4.7 & \negusd{4,722} & +\$5,711 & \negusd{9,939} & \negusd{11,134} & \negnum{7.76} (\negnum{9.31} to \negnum{6.21}) & 54 / 54 & \$63 \\
    Muse Ember & \negusd{5,383} & +\$4,108 & \negusd{10,128} & \negusd{11,795} & \negnum{7.24} (\negnum{9.79} to \negnum{4.69}) & 54 / 54 & \$1 \\
    Muse Spark 1.3 & \negusd{5,669} & +\$4,541 & \negusd{10,774} & \negusd{12,081} & \negnum{7.63} (\negnum{9.97} to \negnum{5.30}) & 54 / 54 & \$31 \\
    GLM 5.3* & \negusd{6,454} & +\$2,250 & \negusd{10,806} & \negusd{12,866} & \negnum{7.85} (\negnum{9.83} to \negnum{5.87}) & 53 / 54 & \$29 \\
    Claude Haiku 5.5 & \negusd{7,440} & +\$480 & \negusd{11,400} & \negusd{13,852} & \negnum{8.48} (\negnum{10.78} to \negnum{6.18}) & 54 / 54 & \$1 \\
    Muse Wonder 1.0 & \negusd{8,291} & +\$2,701 & \negusd{13,786} & \negusd{14,703} & \negnum{9.33} (\negnum{12.12} to \negnum{6.54}) & 54 / 54 & \$42 \\
    GPT-6 Luna & \negusd{8,667} & \negusd{117} & \negusd{12,942} & \negusd{15,079} & \negnum{9.44} (\negnum{11.50} to \negnum{7.39}) & 53 / 54 & \$2 \\
    \bottomrule
  \end{tabular}%
  }
  \vskip 0.18in
  \caption{Controlled shortfall-cover adaptation from Day 14 to Day 88 across all 15 models with cover-trajectory series (Gemini 4 Hydrogen was added after the trajectory export; customer-loss quarters versus each model's own normal-quarter controls; 10,000-resample bootstrap 95\% intervals; zero models have an interval strictly above zero).}
  \label{tab-cover-all}
  \vskip 0.04in
  \centering
  \resizebox{\textwidth}{!}{%
  \begin{tabular}{lcccccc}
    \toprule
    Model & Customer Day 14 $\to$ 88 & Customer Rise (pp) & Normal Day 14 $\to$ 88 & Normal Rise (pp) & Difference (95\% CI, pp) & Pinned at 100\% (Cust / Norm) \\
    \midrule
    GPT-6 Astra & 93.5\% $\to$ 96.2\% & +2.7 & 77.5\% $\to$ 82.5\% & +5.0 & \negnum{2.4} (\negnum{9.7} to +4.2) & 16 / 18, 13 / 18 \\
    Gemini 4 Carbon & 100.0\% $\to$ 100.0\% & 0.0 & 100.0\% $\to$ 100.0\% & 0.0 & 0.0 (0.0 to 0.0) & 18 / 18, 17 / 17 \\
    Claude Fable 5.2 & 29.5\% $\to$ 55.2\% & +25.7 & 19.5\% $\to$ 24.9\% & +5.4 & +20.2 (\negnum{3.8} to +44.2) & 0 / 18, 1 / 18 \\
    Claude Fable 5.1 & 23.3\% $\to$ 36.5\% & +13.2 & 13.4\% $\to$ 39.0\% & +25.6 & \negnum{12.4} (\negnum{34.8} to +9.8) & 2 / 18, 0 / 18 \\
    GPT-6.1 Sol & 52.0\% $\to$ 61.2\% & +9.3 & 52.6\% $\to$ 55.0\% & +2.5 & +6.8 (\negnum{2.8} to +17.9) & 8 / 18, 8 / 18 \\
    Gemini 4 Argon & 100.0\% $\to$ 100.0\% & 0.0 & 100.0\% $\to$ 94.4\% & \negnum{5.6} & +5.6 (0.0 to +16.7) & 18 / 18, 17 / 18 \\
    Claude Opus 5.5 & 0.4\% $\to$ 11.5\% & +11.1 & 6.2\% $\to$ 7.7\% & +1.5 & +9.6 (\negnum{3.0} to +26.3) & 0 / 18, 1 / 18 \\
    DeepSeek V4.1 Flash* & 0.4\% $\to$ 6.3\% & +5.9 & 2.3\% $\to$ 1.4\% & \negnum{0.9} & +6.8 (\negnum{0.3} to +17.2) & 0 / 18, 0 / 18 \\
    Grok 4.7 & 4.1\% $\to$ 1.7\% & \negnum{2.4} & 5.3\% $\to$ 4.7\% & \negnum{0.6} & \negnum{1.9} (\negnum{6.9} to +3.0) & 0 / 18, 0 / 18 \\
    Muse Ember & 2.6\% $\to$ 2.0\% & \negnum{0.6} & 5.3\% $\to$ 3.5\% & \negnum{1.8} & +1.3 (\negnum{1.6} to +4.5) & 0 / 18, 0 / 18 \\
    Muse Spark 1.3 & 1.7\% $\to$ 3.5\% & +1.8 & 0.0\% $\to$ 6.6\% & +6.6 & \negnum{4.8} (\negnum{18.4} to +6.0) & 0 / 18, 0 / 18 \\
    GLM 5.3* & 21.0\% $\to$ 28.4\% & +7.5 & 26.7\% $\to$ 30.5\% & +3.8 & +3.6 (\negnum{23.5} to +31.6) & 2 / 18, 2 / 18 \\
    Claude Haiku 5.5 & 1.4\% $\to$ 1.0\% & \negnum{0.4} & 1.9\% $\to$ 2.2\% & +0.3 & \negnum{0.7} (\negnum{1.8} to 0.0) & 0 / 18, 0 / 18 \\
    Muse Wonder 1.0 & 2.6\% $\to$ 0.7\% & \negnum{1.9} & 1.8\% $\to$ 1.9\% & +0.1 & \negnum{2.0} (\negnum{4.6} to +0.3) & 0 / 18, 0 / 18 \\
    GPT-6 Luna & 9.1\% $\to$ 6.3\% & \negnum{2.8} & 9.3\% $\to$ 6.0\% & \negnum{3.3} & +0.5 (\negnum{6.1} to +8.0) & 0 / 18, 0 / 18 \\
    \bottomrule
  \end{tabular}%
  }
\end{table}

\section{Prompt Instructions}
\label{app-prompts}

Four prompt settings appear in the paper. All 16 leaderboard models ran the \emph{Benchmark Prompt}, which defines the wholesaler's objective, valuation rules, SQL and purchasing tools, and memory limits without prescribing operational formulas or customer-cover policies. For our open-model study, the \emph{Starting Qwen prompt} keeps those benchmark rules and adds Qwen XML tool syntax, and the \emph{Training Prompt} (used during shared-state RL) kept that exact environment and XML tool-syntax header and appended SQL stockout-adjustment math plus qualitative wholesaler advice. The \emph{Evaluation Prompt} (unseen during RL and shared by both the untrained and trained Qwen3.8 models at evaluation) added batching and SQL arithmetic limits alongside two quantitative rules read from Businesses 1--4 records. Below we quote the exact operational clauses across the Training Prompt and Evaluation Prompt.

\subsection{Demand Censoring Adjustment (Training Prompt and Evaluation Prompt)}
Both the Training Prompt and Evaluation Prompt included the illustrative partial-stockout adjustment $\hat{d}_{\mathrm{pos}} = \mathrm{grams\_sold} \times \frac{16}{16 - h_{\mathrm{oos}}}$ and separated walk-in demand estimates from booked account orders.
\begin{tracequote}
Sales are a floor on demand when the shelf ran empty. If a product sold 9,000 g on a day with 8 of 16 selling hours out of stock, walk-in demand that day was plausibly about 9,000 x 16 / 8 = 18,000 g, not 9,000 g.
\end{tracequote}

\subsection{Batching and SQL Arithmetic Limits (Added in Evaluation Prompt)}
Under the Training Prompt, which omitted per-response call limits and SQL arithmetic rules, 42\% (5 of 12) of full-quarter Business 5 runs failed to complete, four on 65,536-token output overflows (Days 33 to 84) and one on ten consecutive tool-contract rejections (Day 71). The Evaluation Prompt added explicit response batching and SQL sum rules.
\begin{tracequote}
Keep each response to about 15 tool calls or fewer and split a larger set of orders across responses, reading the statuses in between.
\end{tracequote}
\begin{tracequote}
Do sums and per-product arithmetic with SQL aggregates rather than long calculations by hand, and keep each morning's reasoning to what changes the decision.
\end{tracequote}

\subsection{Account Protection Rule (Training Prompt Versus Evaluation Prompt)}
In the Training Prompt, the account-protection advice included a conditional cost exception, which left 586~kg of validation shortfalls (513~kg after adding run-safety limits, Table~\ref{tab-prompt-ablations}) because 2.54$\times$ local cover loses money on individual order lines.
\begin{tracequote}
The leading buyers set a standing local-buy instruction (set\_shortfall\_cover with sku\_id '*') for every account on the first morning and kept it unless local prices made it clearly uneconomic.
\end{tracequote}
In the Evaluation Prompt, training-business comparisons replaced that conditional exception with standing Day-0 wildcard cover across all accounts.
\begin{tracequote}
The leading buyers set a standing local-buy instruction (set\_shortfall\_cover with sku\_id '*' and cover true) for every account on the first morning and seldom switched it off; buyers who switched cover on and off product by product to save the local premium did worse, because the premium on a covered line is usually much smaller than the account sales a shorted line loses.
\end{tracequote}

\subsection{Pre-Markdown Lot Cap (Added in Evaluation Prompt)}
Whereas the Training Prompt advised sizing orders to cover censoring-adjusted demand plus a delivery buffer, the Evaluation Prompt bounded usable stock plus expected arrivals within sales before \texttt{markdown\_after\_days}. As analyzed in Section~\ref{sec-coordination}, expired or culled weight was 0.70\% of received weight, with \$488 in spoilage write-offs, while overall Business 6 stock averaged 0.89 days and 70.3\% of emergency spend fell on 2-day retail-markdown items.
\begin{tracequote}
For each product, keep usable stock plus expected arrivals within what you expect to sell (walk-in plus account bookings) before its markdown\_after\_days; for products that mark down within a few days, order smaller lots more often instead of one large lot.
\end{tracequote}

\section{Additional Transcript Excerpts and Campaign Accounting}
\label{app-repro}

\subsection{Additional Rollout and Test Transcripts}
Every transcript and prompt excerpt quoted in the main text and appendices is reproduced verbatim from recorded episode transcripts, buyer notebooks, and system prompts.

\subsubsection{Censored-Demand Calculation on Sealed Business 6}
In a trained customer-loss quarter (seed 0) on Day 15, observing zero potato sales with 16 stockout hours on Day 13 and 7,469~g sold with 2 stockout hours on Day 14, the policy corrected for censoring before ordering a full wholesale case.
\begin{tracequote}
Let me estimate true demand from the sales data (d13, d14, with OOS correction)
\end{tracequote}
\begin{tracequote}
Potatoes d13: 0, 16h; d14: 7,469, 2h OOS → *16/14 = 8,534 → \textasciitilde8k+
\end{tracequote}

\subsubsection{Pre-Markdown Lot Deferral and Unfunded Cover Diagnosis on Business 6}
In a trained normal quarter (seed 3) on Day 13, treating a 2-day retail markdown step on 7-to-10-day green beans as a lot cap led the policy to defer 2-day backup supplier orders; in a supplier-shock quarter (seed 3) on Day 83, the policy diagnosed unfunded cover rows after cash depletion.
\begin{tracequote}
sup\_b is +2 → d15 arrival. d15 demand = lines + pos; can reassess tomorrow morning when d15 lines post
\end{tracequote}
\begin{tracequote}
All the local\_purchases rows I've seen so far are status=skipped\_funds, with amount 0. That means on d80/d81, local cover did nothing
\end{tracequote}

\subsubsection{Supplier Degradation Tracking Alongside One-Day Lock-In}
In a trained supplier-shock quarter (seed 2, notebook and Day 81), the buyer explicitly logged the 1-day vendor's cancellation rate while continuing to rely on its 1-day lead time.
\begin{tracequote}
Suppliers: sup\_c d+1 NET21 (15-20\% pre-cancel/case), sup\_d d+2 NET7 reliable
\end{tracequote}
\begin{tracequote}
Candidate sup\_c (d82 arrival, cheap, but 1-day late risk + 15-20\% cancellation):
\end{tracequote}

\subsubsection{Quarter-Close Tapering and Tool Self-Repair on Business 6}
Across unseen Business 6 notebooks, the trained policy recorded quarter-close tapering and diagnosed idempotency-key reuse on rejected purchase orders.
\begin{tracequote}
d89 plan: NO BUY (d90+ arrivals worthless at close).
\end{tracequote}
\begin{tracequote}
7. TOOL FIX: rejected PO burns its idempotency\_key -> new key on retry.
\end{tracequote}

\subsubsection{Rare-Win Credit and Quarter-Close Tapering in Training Continuations}
On a Day-66 Business 2 customer-loss state against Claude Opus 5.5 (\negusd{3,953} reference profit, 345~kg prior shortfalls), only 1 of 4 continuations (\negusd{3,693}, weighted $+3.0$ by MaxRL) enabled wildcard cover on Day 82 while holding spoilage to \$1,241, saving +\$226 to +\$398 on terminal customer reserve over the other three continuations from the same state. On a Day-80 Business 4 normal state against GPT-6.1 Sol (+\$6,576 reference profit), one continuation achieved the highest operating book gain (+\$7,135) yet fell below the reference profit overall (+\$6,276, $A = -1.0$) after ordering 3,458~kg during Days 85--89 and incurring a \negusd{808} unsold closing-stock write-off versus \negusd{34} and \negusd{50} for the two reference-beating continuations.

\subsection{Campaign Accounting and Experimental Design Summary}
Across all model families and sessions in the post-training research program (spanning early single-decision proxy prototypes, Qwen3.6-35B supervised and RL runs, and Qwen3.8-27B instruction and RL runs), 94 total experiments comprised 62 weight-updating training runs and 32 evaluation configurations (\textasciitilde\$6,500 list-price compute for the full-buyer campaign; \$605.82 for the 4-update flagship MaxRL run). Table~\ref{tab-exp-summary} summarizes the store splits, prompt settings, and experimental configurations across the study, and Table~\ref{tab-individual-runs} lists every individual quarter outcome under the Evaluation Prompt on Business 5 (validation) and Business 6 (sealed test).

\begin{table}[!htbp]
  \caption{Summary of store splits, prompt settings, and training and evaluation configurations.}
  \label{tab-exp-summary}
  \vskip 0.04in
  \centering
  \resizebox{\textwidth}{!}{%
  \begin{tabular}{lll}
    \toprule
    Component & Sample / Scale & Role and Experimental Protocol \\
    \midrule
    Businesses 1--4 (training) & 12 tasks (4 stores $\times$ 3 conditions) & \$91K--\$210K baseline sales; transcripts for SFT and starting states for continuation RL \\
    Business 5 (validation) & 3 tasks (1 store $\times$ 3 conditions) & \$340K baseline sales; prompt refinement and optimizer selection \\
    Business 6 (sealed test) & 3 tasks (1 store $\times$ 3 conditions) & \$109K baseline sales; evaluated once at campaign close \\
    Benchmark Prompt & 16 models, 859 scored runs & Objective, valuation rules, SQL and order tools, memory limits \\
    Starting Qwen prompt & Qwen3.6 / Qwen3.8 baseline & Benchmark rules plus Qwen XML tool-call syntax \\
    Training Prompt & Shared-state RL prompt & Adds SQL stockout-censoring adjustment and wholesaler advice \\
    Evaluation Prompt & Validation and test evaluation & Adds $\le 15$-call batching, SQL sums, Day-0 cover, and pre-markdown cap \\
    Day-60 continuation panel & 24 shared states $\times$ 4 arms & Compares SFT, Qwen3.6/3.8 backbone, and Training Prompt from identical prefixes \\
    Shared-state binary MaxRL & 4 updates $\times$ 16 continuations & 4 states $\times$ 4 rollouts/step from Days 60--85, lr $3\times 10^{-5}$, temp $1.0$, CISPO ($c_{\max}=4.0$), \$605.82 \\
    Continuous-dollar RLOO & 6 updates $\times$ 16 continuations & Updates 0--3 on identical states, updates 4--5 from Days 35--59, lr $3\times 10^{-5}$, temp $1.0$, \$1,229.56 \\
    Held-out Business 6 pair & 12 quarters per model & 4 seeds $\times$ 3 conditions at temp $1.0$ comparing trained vs.\ untrained under Evaluation Prompt \\
    \bottomrule
  \end{tabular}%
  }
\end{table}

\begin{table}[!htbp]
  \caption{Individual quarter outcomes under the Evaluation Prompt across all 24 validation runs (Business 5) and all 24 sealed-test runs (Business 6). Completed runs report graded business return; the single early model termination (Business 5 supplier shock, Seed 3, 10 consecutive tool-contract errors on Day 75) has no completed-quarter grade and receives the pre-declared \negusd{25,000} failure score in all-run means.}
  \label{tab-individual-runs}
  \vskip 0.04in
  \centering
  \resizebox{\textwidth}{!}{%
  \begin{tabular}{llcccccc}
    \toprule
    & & \multicolumn{2}{c}{Business 5 (Validation)} & \multicolumn{4}{c}{Business 6 (Sealed Test)} \\
    \cmidrule(lr){3-4} \cmidrule(lr){5-8}
    Condition & Seed & Untrained & Trained & Untrained & Untrained Shorts & Trained & Trained Shorts \\
    \midrule
    Normal & 0 & +\$8,677 & +\$7,189 & \negusd{6,036} & 13 kg & \negusd{3,576} & 3 kg \\
    Normal & 1 & +\$3,401 & +\$1,263 & \negusd{4,675} & 2 kg & \negusd{1,518} & 0 kg \\
    Normal & 2 & +\$3,281 & +\$7,920 & \negusd{4,807} & 30 kg & \negusd{4,625} & 0 kg \\
    Normal & 3 & +\$29 & +\$2,563 & \negusd{3,474} & 12 kg & \negusd{4,509} & 0 kg \\
    \midrule
    Customer loss & 0 & \negusd{536} & +\$1,805 & \negusd{1,373} & 0 kg & \negusd{2,891} & 0 kg \\
    Customer loss & 1 & +\$6,910 & +\$2,073 & \negusd{3,611} & 0 kg & \negusd{1,407} & 0 kg \\
    Customer loss & 2 & +\$9,207 & +\$7,116 & \negusd{1,842} & 0 kg & +\$984 & 0 kg \\
    Customer loss & 3 & +\$9,375 & +\$4,885 & \negusd{3,739} & 0 kg & \negusd{2,997} & 7 kg \\
    \midrule
    Supplier shock & 0 & +\$6,913 & \negusd{733} & \negusd{13,913} & 441 kg & \negusd{3,182} & 0 kg \\
    Supplier shock & 1 & \negusd{2,007} & \negusd{1,754} & \negusd{3,563} & 34 kg & \negusd{9,588} & 154 kg \\
    Supplier shock & 2 & \negusd{5,655} & +\$552 & \negusd{12,907} & 223 kg & \negusd{7,979} & 112 kg \\
    Supplier shock & 3 & Crash d75 (\negusd{25,000}) & \negusd{3,844} & \negusd{11,872} & 142 kg & \negusd{12,657} & 420 kg \\
    \midrule
    All-run mean (with failure score) & 12 / 12 & +\$1,216 (11/12 valid) & +\$2,419 (12/12) & \negusd{5,984} (12/12) & 75 kg & \negusd{4,495} (12/12) & 58 kg \\
    Completed-only mean (equal condition wt.) & Valid only & +\$3,279 (11/12 valid) & +\$2,419 (12/12) & \negusd{5,984} (12/12) & 75 kg & \negusd{4,495} (12/12) & 58 kg \\
    \bottomrule
  \end{tabular}%
  }
\end{table}

\end{document}